%% file: natural_language_reasoning_nli.tex
\documentclass[11pt,a4paper]{article}
\usepackage[hyperref]{styles/acl2020}
\usepackage{times}
\usepackage{latexsym}
\renewcommand{\UrlFont}{\ttfamily\small}

\usepackage{microtype}

\usepackage{graphicx}
\usepackage{color}
\usepackage{amsmath}
\usepackage{multirow}
\usepackage{enumitem}

\aclfinalcopy 
\def\aclpaperid{2966} 

\input{natural_language_reasoning_nli_defs}

\title{NILE : Natural Language Inference with Faithful Natural Language Explanations}

\author{Sawan Kumar \\
  Indian Institute of Science, Bangalore \\
  \texttt{sawankumar@iisc.ac.in} \\\And
  Partha Talukdar \\
  Indian Institute of Science, Bangalore \\
  \texttt{ppt@iisc.ac.in} \\}

\date{}

\begin{document}
\maketitle

\begin{abstract}

The recent growth in the popularity and success of deep learning models on NLP classification tasks has accompanied the need for generating some form of  natural language explanation of the predicted labels. Such generated natural language (NL) explanations are expected to be \emph{faithful}, i.e., they should correlate well with the model's internal  decision making. In this work, we focus on the task of natural language inference (NLI) and address the following question: \emph{can we build NLI systems which produce labels with high accuracy, while also generating faithful explanations of its decisions?} We propose \systemfull{} (\system{}), a novel NLI method which utilizes auto-generated label-specific NL explanations to produce labels along with its faithful explanation. We demonstrate \system{}'s effectiveness over previously reported methods through automated and human evaluation of the produced labels and explanations. Our evaluation of \system{} also supports the claim that accurate systems capable of providing testable explanations of their decisions can be designed. We discuss the faithfulness of \system{}'s explanations in terms of sensitivity of the decisions to the corresponding explanations. We argue that explicit evaluation of faithfulness, in addition to label and explanation accuracy, is an important step in evaluating model's explanations. Further, we demonstrate that task-specific probes are necessary to establish  such sensitivity.
\end{abstract}

\input{sections/intro}

\input{sections/related}
\input{sections/background}
\input{sections/method}
\input{sections/experiments}
\input{sections/conclusion}

\section*{Acknowledgments}

We thank the anonymous reviewers for their constructive comments. This work is supported by the Ministry of Human Resource Development (Government of India). We would also like to thank HuggingFace for providing a state-of-the-art Transformers library for natural language understanding. Finally, we want to thank the annotators who annotated generated explanations for correctness.

\bibliography{natural_language_reasoning_nli}
\bibliographystyle{styles/acl_natbib}

\appendix
\input{sections/appendix}

\end{document}

%% file: natural_language_reasoning_nli_defs.tex
\newcommand{\refeqn}[1]{Equation~\ref{#1}}
\newcommand{\reffig}[1]{Figure~\ref{#1}}
\newcommand{\reftbl}[1]{Table~\ref{#1}}
\newcommand{\refsec}[1]{Section~\ref{#1}}

\newcommand{\reminder}[1]{}

\newcommand{\system}{NILE}
\newcommand{\nophsystem}{NILE-PH}
\newcommand{\nonssystem}{NILE-NS}

\newcommand{\systemfull}{Natural-language Inference over Label-specific Explanations}

\newcommand{\egenerator}{Candidate Explanation Generator}
\newcommand{\reasoner}{Explanation Processor}

\newcommand{\conjecture}{explanation}

\newcommand{\independent}{Independent}
\newcommand{\aggregate}{Aggregate}
\newcommand{\append}{Append}

\usepackage{array}
\newcolumntype{L}[1]{>{\raggedright\let\newline\\\arraybackslash\hspace{0pt}}m{#1}}
\newcolumntype{C}[1]{>{\centering\let\newline\\\arraybackslash\hspace{0pt}}m{#1}}
\newcolumntype{R}[1]{>{\raggedleft\let\newline\\\arraybackslash\hspace{0pt}}m{#1}}

%% file: sections/intro.tex
\section{Introduction}
\label{sec:intro}

Deep learning methods have been employed to improve performance on several benchmark classification tasks in NLP \citep{wang2018glue,wang2019superglue}. Typically, these models aim at improving label accuracy, while it is often desirable to also produce explanations for these decisions \citep{lipton2016mythos,chakraborty2017interpretability}. In this work, we focus on producing natural language explanations for Natural Language Inference (NLI), without sacrificing much on label accuracy.

There has been growing interest in producing natural language explanations for deep learning systems \citep{huk2018multimodal,kim2018textual,ling-etal-2017-program},  including NLI \citep{camburu2018snli}. In general, the explanations from these methods can typically be categorized as post-hoc explanations \citep{lipton2016mythos}. \citet{camburu2018snli} propose an NLI system which first produces an explanation and then processes the explanation to produce the final label. We argue that these explanations also resemble post-hoc explanations (\refsec{sec:motivation}). Further, existing methods don't provide a natural way to test the \textit{faithfulness} of the generated explanations, i.e., how well do the provided explanations correlate with the model's decision making.

\begin{figure*} [thb]
  \centering
  \includegraphics[width=\linewidth]{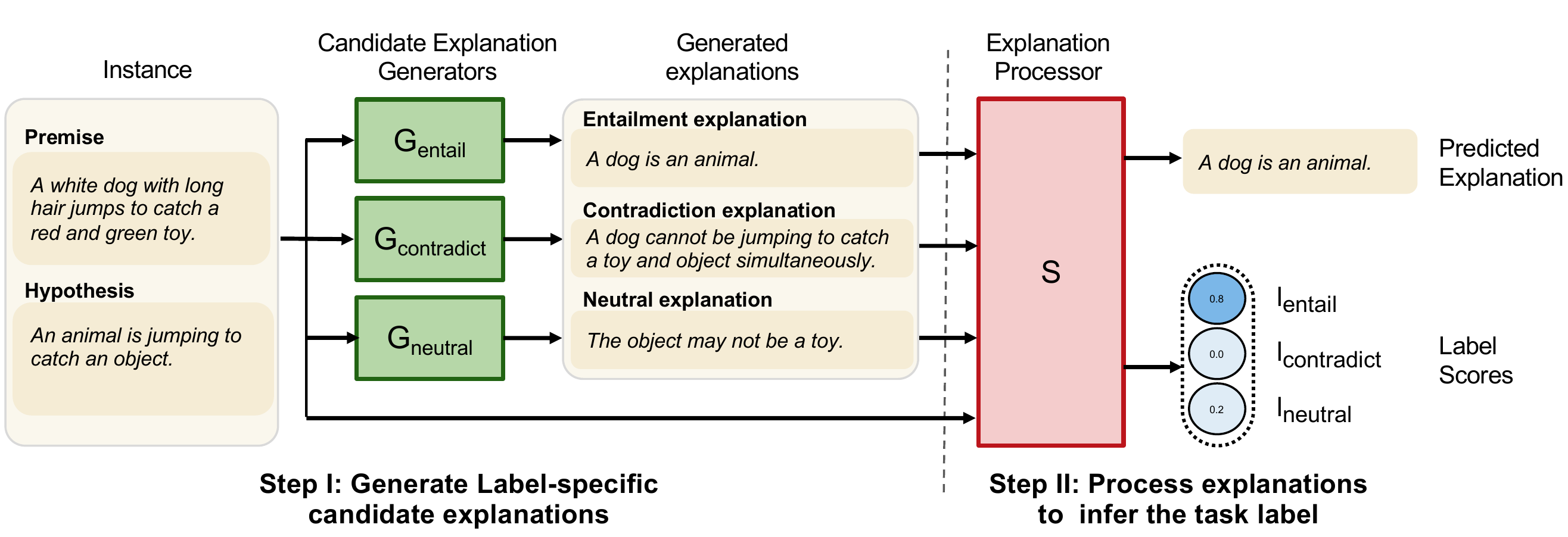}
  \caption{\label{fig:f1_schematic} \textit{Overview of \system{}}: A Premise and Hypothesis pair is input to label-specific \egenerator{}s $G$ which generate natural language explanations supporting the corresponding label. The generated explanations are then fed to the \reasoner{} $S$, which generates label scores using the evidence present in these explanations (see \reffig{fig:f3_explanation_selector} for the architectures used in this work). In addition to the explanations, \system{} also utilizes the premise and hypothesis pair (See \refsec{subsec:using_conjectures_and_examples} for a discussion on the challenges in building such a system). Please see \refsec{sec:method} for details.}
\end{figure*}

We therefore propose \systemfull{} (\system{})\footnote{\system{} source code available at \\ \UrlFont{https://github.com/SawanKumar28/nile}}, which we train and evaluate on English language examples.  Through \system{}, we aim to answer the following question: \\\\
\textit{Can we build NLI systems which produce faithful natural language explanations of predicted labels, while maintaining high accuracy?}\\\\
 Briefly, in \system{}, we first generate natural language \conjecture{}s for each possible decision, and subsequently process these \conjecture{}s to produce the final decision. We argue that such a system provides a natural way of explaining its decisions. The key advantage is the testability of these explanations, in themselves, as well as in terms of the sensitivity of the system's prediction to these explanations. 

We choose NLI due to its importance as an NLP task, and the availability of e-SNLI, a large dataset annotated both with entailment relation labels and natural language human explanations of those labels \citep{camburu2018snli,bowman-etal-2015-large}.

In summary, we make the following contributions in this work.

\begin{enumerate} [noitemsep,nolistsep]
	\item We propose \system{}, an NLI system which generates and processes label-specific \conjecture{}s to infer the task label, naturally providing explanations for its decisions.
	\item We demonstrate the effectiveness of \system{} compared to existing systems, in terms of label and explanation accuracy.
	\item Through \system{}, we provide a framework for generating falsifiable explanations. We propose ways to evaluate and improve the faithfulness of the system's predictions to the generated explanations. We claim that task-specific probes of sensitivity are crucial for such evaluation.
\end{enumerate}

We have released the source code of \system{} to aid reproducibility of the results.

%% file: sections/related.tex
\section{Related Work}
\label{sec:related}

Explainability of a model's predictions has been studied from different perspectives, including feature importance based explanations \citep{ribeiro2016should,lundberg2017unified,chen2018learning}, or post-hoc natural language explanations \citep{huk2018multimodal,kim2018textual,ling-etal-2017-program}. \citet{hendricks2018generating} produce counterfactual natural language explanations for image classification given an image and a counter-class label. \citet{camburu2018snli} propose a model for NLI to first generate a free-form natural language explanation and then infer the label from the explanation. However, as noted by \citet{oana2019make}, the system tends to generate inconsistent explanations. We reason that requiring a model to generate an explanation of the correct output requires it to first infer the output, and the system thus resembles post-hoc explanation generation methods.

Given the diversity of desiderata and techniques for interpretability, the need for understanding interpretation methods and evaluating them has grown. Difficulty in building interpretation models and the lack of robustness of the same are some of the major issues in existing deep neural networks systems \cite{feng-etal-2018-pathologies,ghorbani2019interpretation,oana2019can}. Given these observations, measuring faithfulness, i.e., how well do the provided explanations correlate with the model's decision making, is crucial. \citet{deyoung2019eraser} propose metrics to evaluate such faithfulness of rationales (supporting evidence) for NLP tasks.

Through \system{}, we propose a framework for generating faithful natural language explanations by requiring the model to condition on generated natural language explanations. The idea of using natural language strings as a latent space has been explored to capture compositional task structure \citep{andreas-etal-2018-learning}. \citet{wu-etal-2019-generating} explore improving visual question answering by learning to generate question-relevant captions. \citet{rajani-etal-2019-explain} aim to improve commonsense question answering by first generating commonsense explanations for multiple-choice questions, where the question and the choices are provided as the prompt. Similar to \cite{camburu2018snli}, they learn by trying to generate human-provided explanations and subsequently conditioning on the generated explanation. In \system{}, we instead aim to produce an explanation for each possible label and subsequently condition on the generated label-specific explanations to produce the final decision. 

%% file: sections/background.tex
\section{Background}
\label{sec:background}

In this section, we discuss the datasets (\refsec{subsec:background_data}) and pre-trained models (\refsec{subsec:background_models}) used to build \system{}.

\subsection{Data}
\label{subsec:background_data}

\paragraph{SNLI:} The Stanford NLI dataset \citep{bowman-etal-2015-large} contains samples of premise and hypothesis pairs with human annotations, using Amazon Mechanical Turk. The premises were obtained from pre-existing crowdsourced corpus of image captions. The hypotheses were obtained by presenting workers with a premise and asking for a hypothesis for each label (entailment, neutral and contradiction), resulting in a balanced set of  $\sim$570K pairs.

\paragraph{e-SNLI:} \citet{camburu2018snli} extend the SNLI dataset with natural language explanations of the ground truth labels. The explanations were crowd-sourced using Amazon Mechanical Turk. Annotators were first asked to highlight words in the premise and hypothesis pairs which could explain the labels. Next, they were asked to write a natural language explanation using the highlighted words.

Similar to \citet{camburu2018snli}, for all our experiments, we filter out non-informative examples where the explanations contain the entire text of the premise or hypothesis. In particular, we drop any training example where the uncased premise or hypothesis text appears entirely in the uncased explanation. This leads to a training data size of $\sim$532K examples.

\subsection{Pretrained Language Models}
\label{subsec:background_models}
Transformer architectures \citep{vaswani2017attention} pre-trained on large corpora with self-supervision have shown significant improvements on various NLP benchmarks \citep{devlin-etal-2019-bert,radford2019language,yang2019xlnet,liu2019roberta,lan2019albert}. Improvements have been demonstrated for text classification as well as text generation tasks \citep{lewis2019bart,raffel2019exploring}. In this work, we leverage the implementation of transformer architectures and pre-trained models provided by \citet{Wolf2019HuggingFacesTS}.

\paragraph{GPT-2:} We use the GPT-2 architecture \citep{radford2019language}, which is trained using a causal language modeling loss (CLM), and includes a left-to-right decoder suitable for text generation. In particular, we use the gpt2-medium model. This model has 24 layers, 16 attention heads and a hidden size of 1024 ($\sim$345M parameters). For text generation, the model can be finetuned using CLM on desired text sequences.

\paragraph{RoBERTa:} For classification modules, we leverage RoBERTa \citep{liu2019roberta}, which is trained using a masked language modeling loss (MLM). In particular, we use the roberta-base model. This model has 12 layers, 12 attention heads and a hidden size of 768 ($\sim$125M parameters). For downstream classifications tasks, a classification layer is added over the hidden-state of the first token in the last layer.

%% file: sections/method.tex
\section{\systemfull{} (\system{})}
\label{sec:method}

The overall architecture employed in \system{} is shown in \reffig{fig:f1_schematic}. We introduce the notation used in this paper in \refsec{sec:notation}. We then discuss the motivation for the major design choices in \refsec{sec:motivation}.

\system{} performs the following steps to produce labels and explanations:
\begin{enumerate}[]
\item \textbf{\egenerator{}s:} Label-specific \egenerator{}s first generate \conjecture{}s supporting the respective labels (\refsec{sec:conjecture_generators}).
\item \textbf{\reasoner{}:} The \reasoner{} takes the \conjecture{}s and also the premise and hypothesis pairs as input to produce the task label (\refsec{sec:nli_reasoner}). We also build \nophsystem{}, where the \reasoner{} has access only to the generated explanations (\refsec{subsec:using_ony_conjectures}).
\end{enumerate}

We note that \nophsystem{} more naturally fits the desiderata described in \refsec{sec:intro}, while we design and evaluate \system{} for the more general case where the \reasoner{} also accesses the premise and hypothesis pair. 

In \refsec{sec:baselines}, we describe comparable baseline architectures.

\subsection{Notation}
\label{sec:notation}

We denote each data point by ($p$, $h$), where $p$ is the premise and $h$ the hypothesis sentence. $G$ denotes a model trained to generate natural language explanations. Specifically, $G_x$ denotes a model which generates natural language explanations $t_x$ of type $x$, where $x \in $ \{entail, contradict, neutral\}. We denote the human-provided gold explanation for the correct predictions as $t_g$.
$S$ denotes a module which predicts label scores. The true label for an example is denoted by $y$, while a model prediction is denoted by $y'$, and label scores by $l_x$.

\begin{figure} [thb]
  \centering
  \includegraphics[width=\linewidth]{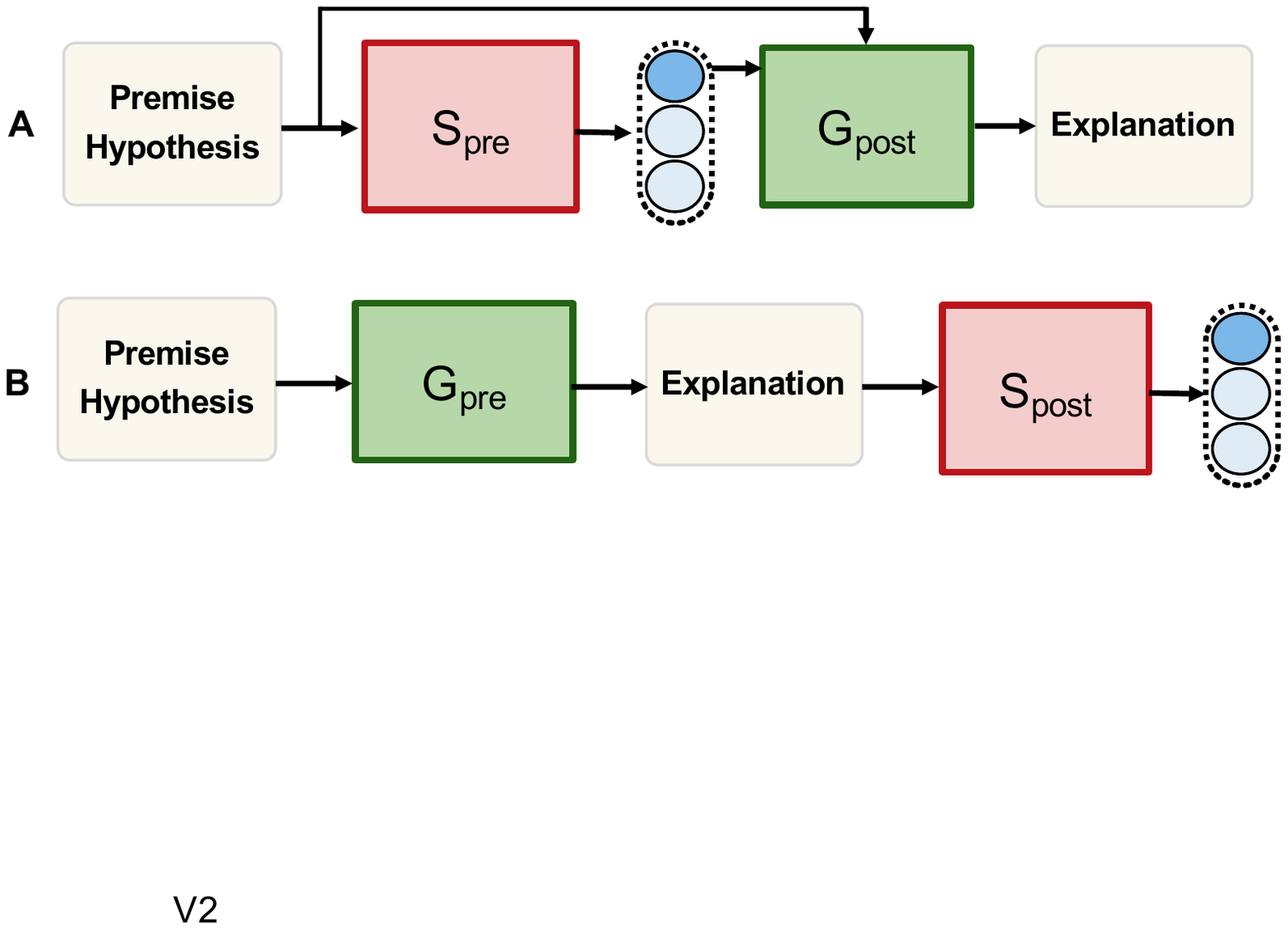}
  \caption{\label{fig:f2_alternatives} \textit{Existing alternative architectures.}: \textbf{A.} \textit{Post-hoc generation}: Given an input instance, first the label is predicted and then an explanation generated conditioned on the label and the input text. \textbf{B.} \textit{ExplainThenPredict} \citep{camburu2018snli}: Given the input instance, first the desired explanation is generated, and then the label is predicted using only the generated explanation. We argue that neither architecture provides a natural way to test the sensitivity of the model's predictions to the generated explanation. Please see \refsec{sec:motivation} for details.}
\end{figure}

\subsection{Why do it this way?}
\label{sec:motivation}

In this section, we describe the motivation for adopting a two-step pipelined approach.

\paragraph{Label-specific explanations:} Consider two alternative existing architectures in \reffig{fig:f2_alternatives}. In \reffig{fig:f2_alternatives}A, a model $S_\text{pre}$ is trained directly on the example sentences ($p$ \& $h$) to produce a label ($y'$), which together with the example sentences are used to produce an explanation $t_g'$ using $G_\text{post}$. It can be argued that while the target explanations may regularize the system, there is no reason for  $t_g'$ to be aligned with the reason why the model chose a particular label.

\reffig{fig:f2_alternatives}B corresponds to a model which has also been trained on e-SNLI \citep{camburu2018snli}. $G_\text{pre}$ is first trained to produce natural language explanations $t_g'$ using human-provided explanations ($t_g$) as targets, using only the example sentences as inputs. A model $S_\text{post}$ then chooses the label corresponding to the generated explanation  $t_g'$. While at first, it appears that this system may provide faithful explanations of its decisions, i.e., the generated explanations are the reason for the label prediction, we argue that it may not be so.

In \reffig{fig:f2_alternatives}B, $G_\text{pre}$ is required to generate the explanation of the correct label for an example. It must first infer that label and then produce the corresponding explanation. Further analysis of the free-form human-provided explanations has revealed clear differences in the form of explanations, through alignment to label-specific templates \citep{camburu2018snli,oana2019make}. The \reasoner{} $S_\text{post}$ then only needs to infer the form of $t_g'$. $G_\text{pre}$ then resembles post-hoc generation methods, with the label (as the form of $t_g'$) and explanation $t_g'$ being produced jointly. The claim is supported by inconsistencies found in the generated explanations \citep{oana2019make}.

Neither architecture allows a natural way to test the sensitivity of the model's predictions to its explanations. In \system{}, we first allow \conjecture{}s for each label, and then require the \reasoner{} to select the correct explanation. This allows us to naturally test whether the model's predictions are indeed due to the selected explanation. This can be done, for example, by perturbing the input to the \reasoner{}.

\paragraph{A pipelined approach:} We use a pipelined approach in \system{} (\reffig{fig:f1_schematic}). The \egenerator{}s are first trained using human-provided explanations. The \reasoner{} takes as input the generated label-specific explanations.  This prevents the system from producing degenerate explanations to aid task performance. It also allows perturbing the generated explanations to probe the system in a more natural way compared to an unintelligible intermediate state of a learnt model. We believe that systems can be designed to work in this setting without compromising task performance.

\begin{figure*} [thb]
  \centering
  \includegraphics[width=\linewidth]{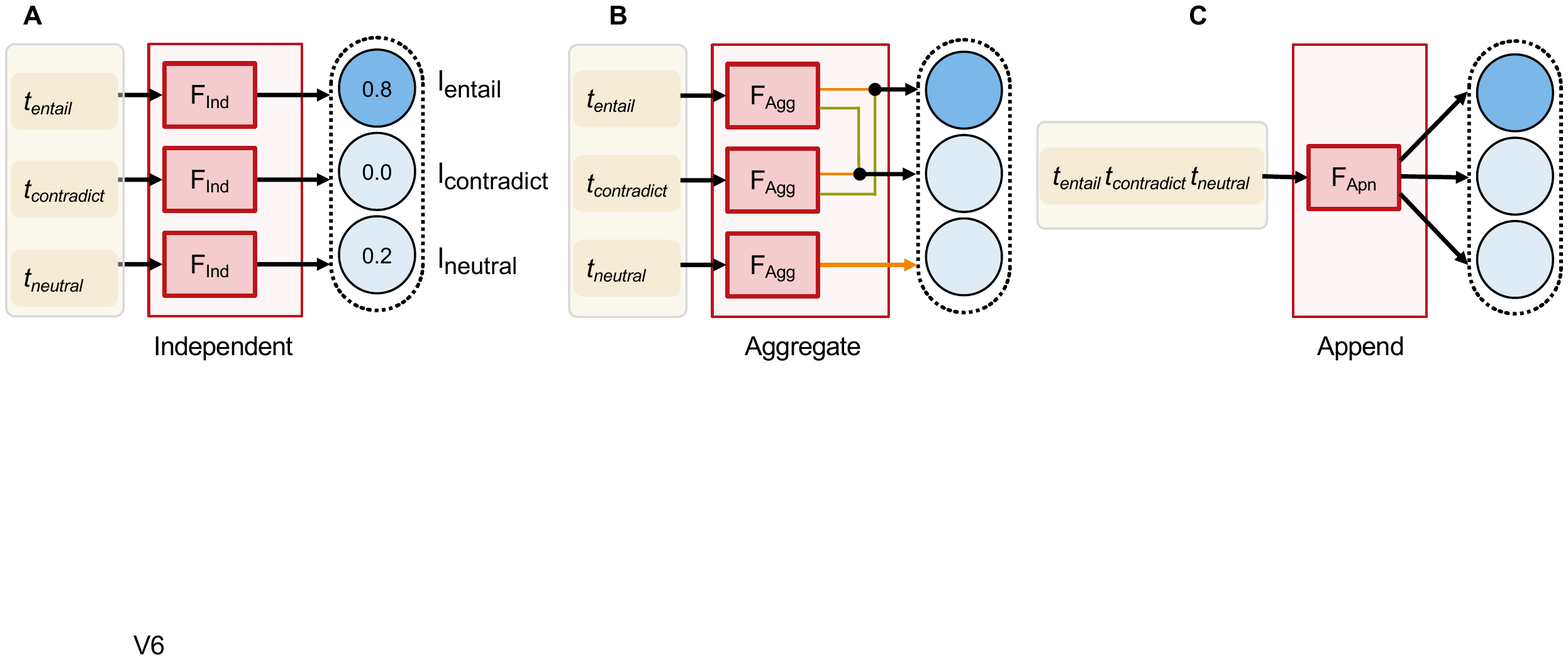}
  \caption{\label{fig:f3_explanation_selector} \textit{\reasoner{} architectures.} \textbf{A.} \independent{} (Ind) collects evidence for a label symmetrically from the corresponding \conjecture{}. \textbf{B.} \aggregate{} (Agg) allows handling missing \conjecture{}s by looking for contradictory evidence. \textbf{C.} \append{} (Apn) allows arbitrary evidence collection for each label. Please see \refsec{subsec:using_ony_conjectures} for details. Premise and hypothesis sentences are processed by additionally providing them to each block $F_z$ where $z \in $ \{Ind, Agg, Apn\}. Please see \refsec{subsec:using_conjectures_and_examples} for details.}
\end{figure*}

\subsection{\egenerator{}s}
\label{sec:conjecture_generators}

We train label-specific \conjecture{} generators, $G_x$, $x \in $ \{entail, contradict, neutral\},  using human-provided explanations of examples with the corresponding label. For example, to train $G_\text{entail}$, we collect all triplets ($p$, $h$, $t_g$) annotated as entailment. We create text sequences of the form: ``\textit{Premise: $p$ Hypothesis: $h$ [EXP] $t_g$ [EOS]}" to fine-tune a pre-trained language model, where \textit{[EXP]} and \textit{[EOS]} are special tokens added to the vocabulary. During fine-tuning, the language modeling loss function is used only over the explanation tokens.

Next, we create prompts of the form  ``\textit{Premise: $p$ Hypothesis: $h$ [EXP]}" and require each trained language model to independently complete the sequence. In this way we obtain label specific \conjecture{}s $t_x$, $t_x = G_x(p,h)$, for $x \in $ \{entail, contradict, neutral\}.

\subsection{\reasoner{}}
\label{sec:nli_reasoner}

The \reasoner{} in \system{} takes as input the generated label-specific \conjecture{}s, as well as  the premise and hypothesis pair to generate label scores $l_x$, $x \in $ \{entail, contradict, neutral\}. During training, these scores are passed through a softmax layer and a cross-entropy loss is used to generate the training signal. During testing, the label with the maximum score is selected.

We leverage a pre-trained roberta-base model for all our experiments, and fine-tune it as specified in the following subsections. In each case, any intermediate scores are generated through transformations of the first token (\texttt{[CLS]}) embedding from the last layer.
We define:
\begin{equation*}
\label{eqn:generation}
	F_{\mathrm{model}}(\texttt{inp}) =\mathrm{tanh}(W . \mathrm{CLS}_{\mathrm{embed}}(\texttt{inp}))
\end{equation*}
where \texttt{inp} is a pair of sequences in \system{}, a single sequence in \nophsystem{}, and $W$ are the learnable parameters for the model.

For simplicity, and to elucidate the desired behavior, we first describe how \conjecture{}s are processed in \nophsystem{} (\refsec{subsec:using_ony_conjectures}).  We then discuss the construction of \system{}, a potential issue, and a fix for the same (\refsec{subsec:using_conjectures_and_examples}).

\subsubsection{Processing Explanations}
\label{subsec:using_ony_conjectures}
In this section, we describe how explanations are processed in \nophsystem{}, which is generalized in \system{} (\refsec{subsec:using_conjectures_and_examples}).
We experiment with three architectures, described below (also see \reffig{fig:f3_explanation_selector}).

\paragraph{A. \independent{}:} In the \independent{} model, \conjecture{}s are fed to $F_{\text{Ind}}$,  which generates a score for each \conjecture{}s independently:
\begin{equation}
\label{eqn:independent}
	l_{x}  = W_\text{Ind} F_{\text{Ind}}(t_x) 
\end{equation}
where $x \in $ \{entail, contradict, neutral\}. We expect this score to represent the truthfulness of the input \conjecture{}.

\paragraph{B. \aggregate{}:} The \independent{} model would need all three \conjecture{}s to be available to reliably produce label scores. We believe a system should be able to handle one or more missing or ambiguous \conjecture{}s. For example, the entailment \conjecture{}: ``$t_\text{entail}$:\textit{ A dog is a cat}" would provide evidence for contradiction. To capture this notion, we require the \reasoner{} to produce two intermediate scores $V_1$ and $V_2$, where we expect $V_1$ to collect evidence supporting an input claim and $V_2$ to collect evidence against an input claim:
\begin{equation}
\label{eqn:aggregate_intermediate}
	V_i(x) = W_{\text{Agg},i} F_{\text{Agg}}(t_x), \:\text{where}\: i  \in \{ 1, 2\}
\end{equation}

The intermediate score are then aggregated into the final label scores:
\begin{align}
\label{eqn:aggregate}
\begin{split}
	l_\text{entail}   &= \mathrm{Cmb}(V_1(t_\text{entail}), V_2(t_\text{contradict})) \\
	l_\text{contradict}   &= \mathrm{Cmb}(V_1(t_\text{contradict}), V_2(t_\text{entail})) \\
	l_\text{neutral}   &= V_1(t_\text{neutral})
\end{split}
\end{align}
where $\mathrm{Cmb}$ is the \texttt{LogSumExp} function.
The reason for this choice of aggregation is that while evidence against entailment might point to contradiction and vice versa, evidence against neutral doesn't necessarily provide any information about entailment or contradiction relations.

\paragraph{C. \append{}:} Finally, to allow the model to reason arbitrarily between the three generated \conjecture{}s, we created a single sequence, $\mathrm{concat}_{ecn}$: ``\textit{entailment: $t_\text{entail}$ contradiction: $t_\text{contradict}$ neutral: $t_\text{neutral}$}", and generate the scores as follows:
\begin{equation}
\label{eqn:append}
	l_x   = W_{\text{Apn},x} F_{\text{Apn}}(\mathrm{concat}_{ecn})
\end{equation}
where $x \in $ \{entail, contradict, neutral\}.

\subsubsection{Processing Premise and Hypothesis}
\label{subsec:using_conjectures_and_examples}
In \system{}, to process premise $p$ and hypothesis $h$, we first concatenate $p$ and $h$ into $\mathrm{concat}_{ph}$: ``\textit{Premise: p Hypothesis: h}". The label scores are then obtained as in \refsec{subsec:using_ony_conjectures}, by modifying \refeqn{eqn:independent}, \ref{eqn:aggregate_intermediate} and \ref{eqn:append} as follows: replace $F_z(x)$ by $F_z(\mathrm{concat}_{ph}, x)$, where $z \in $ \{Ind, Agg, Apn\}.
We note that appending the example sentences to the generated \conjecture{}s (as in \append{}) would result in having no control over whether the \conjecture{}s are used for the final prediction. The case for \independent{} and \aggregate{} is not immediately clear. We now discuss a potential issue with these architectures when processing premise and hypothesis text, and suggest a fix for the same.

\begin{table*}[thb]
\small
\centering
\begin{tabular}{|l|l|c|c|c|c|c|c|c|}
\hline
\multicolumn{2}{|c|}{\multirow{3}{*}{Model}} &
  \begin{tabular}[c]{@{}c@{}}SNLI\\ Dev\end{tabular} &
  \begin{tabular}[c]{@{}c@{}}SNLI\\ Test\end{tabular} &
  \multicolumn{5}{c|}{\begin{tabular}[c]{@{}c@{}}Explanation evaluation on \\ first 100 SNLI Test Samples\end{tabular}} \\ \cline{3-9} 
\multicolumn{2}{|c|}{} &
  \multirow{2}{*}{\begin{tabular}[c]{@{}c@{}}\\ Label\\ Accuracy\end{tabular}} &
  \multirow{2}{*}{\begin{tabular}[c]{@{}c@{}}\\ Label\\ Accuracy\end{tabular}} &
  \multirow{2}{*}{\begin{tabular}[c]{@{}c@{}} A:\\ Correct\\ Labels\end{tabular}} &
  \multicolumn{2}{c|}{\begin{tabular}[c]{@{}c@{}}Averaged \\ over annotators\end{tabular}} &
  \multicolumn{2}{c|}{\begin{tabular}[c]{@{}c@{}}Annotators \\ in-agreement\end{tabular}} \\ \cline{6-9} 
\multicolumn{2}{|c|}{} &
   &
   &
   &
  \begin{tabular}[c]{@{}c@{}}B: \\ Correct\\ Expl.\end{tabular} &
  B/A &
  \begin{tabular}[c]{@{}c@{}}C:  \\ Correct\\ Expl.\end{tabular} &
  C/A \\ \hline \hline
\multicolumn{2}{|l|}{SemBERT$^\#$ \citep{zhang2019semantics}}                                            & 92.2  & 91.9  & -  & -    & -     & -  & -     \\ \hline \hline
\multirow{2}{*}{\begin{tabular}[c]{@{}l@{}}ETPA\\ \citep{camburu2018snli} \end{tabular}} & Reported    & 81.71 &       & -  & -    & 64.27 & -  & -     \\ \cline{2-9} 
                                                                     & Reproduced  & 86.98 & 86.22 & 77 & 71.2 & \textbf{92.47} & 59 & 76.62 \\ \hline \hline
\multicolumn{2}{|l|}{NILE:post-hoc}                                                & \textbf{91.86} & \textbf{91.49} & 90 & 81.4 & 90.44 & 68 & 75.56 \\ \hline
\multirow{3}{*}{NILE-PH}                                             & Independent & 84.69 & 84.13 & 78 & 72.0 & 92.31 & 61 & 78.21 \\ \cline{2-9} 
                                                                     & Aggregate   & 85.71 & 85.29 & 80 & 73.4 & 91.75 & 62 & 77.50 \\ \cline{2-9} 
                                                                     & Append      & 88.49 & 88.11 & 85 & 78.0 & 91.76 & 66 & 77.65 \\ \hline \hline
\multirow{3}{*}{NILE-NS}                                             & Independent & 91.56 & 90.91 & 88 & 80.8 & 91.82 & \textbf{69} & \textbf{78.41} \\ \cline{2-9} 
                                                                     & Aggregate   & 91.55 & 91.08 & 89 & 80.6 & 90.56 & 68 & 76.40 \\ \cline{2-9} 
                                                                     & Append      & 91.74 & 91.12 & 89 & 80.4 & 90.34 & 67 & 75.28 \\ \hline \hline
\multirow{2}{*}{NILE}                                                & Independent & 91.29 & 90.73 & \textbf{91} & \textbf{82.4} & 90.55 & \textbf{69} & 75.82 \\ \cline{2-9} 
                                                                     & Aggregate   & 91.19 & 90.91 & 90 & 81.4 & 90.44 & 68 & 75.56 \\ \hline
\end{tabular}
\caption{\label{tbl:label_accuracy} \textit{Comparison of label and explanation accuracy on the \textbf{in-domain} SNLI evaluation sets.} Models are selected using the Dev set label accuracy over 5 runs with different seeds of random initialization. Mean (and standard deviation) over the 5 runs are reported in the Appendix. $^\#$ indicates the best reported result at {\UrlFont{https://nlp.stanford.edu/projects/snli/}} at the time of writing. Note that SemBERT does not provide natural language explanations and is reported here only for reference. Bold numbers indicate highest among methods that produce explanations. Explanations are evaluated on the first 100 SNLI Test examples. We present reported  numbers of ETPA \citep{camburu2018snli} as well as the results with our reproduction of ETPA. ETPA (reproduced) is directly comparable with \system{} (\refsec{sec:baselines}). \nophsystem{} competes with or outperforms ETPA baselines on label accuracy, while \nonssystem{}  and \system{} provide significant gains in label accuracy. \system{} and \nonssystem{} are competitive with the best reported results in terms of label accuracies. We report the number of correct explanations, averaged across annotators (B) as well as when all annotators agree on correctness (C). All \system{} variants are able to provide more correct explanations than the ETPA baseline. We also report the percentage of correct explanations in the subset of correct label predictions (B/A, C/A). On this metric, \system{} variants are comparable with the ETPA baseline. However, the real value of \system{} lies in being able to probe the faithfulness of its decisions (\refsec{sec:sensitivity_analysis}). Further, \system{} explanations generalize significantly better on out-of-domain examples (See \reftbl{tbl:label_accuracy_mnli}). Please see \refsec{sec:overall_results} for details.}
\end{table*}

\paragraph{The issue:} We expect \system{} to answer the question: Is ($\mathrm{concat}_{ph},t_x$), where $x \in $ \{entail, contradict, neutral\}, a valid instance-explanation pair? The \independent{} and \aggregate{} architectures for \system{} have been designed such that the model can't ignore the label-specific \conjecture{}s. For example, the \independent{} model will produce identical scores for each output label, if it chooses to completely ignore the input \conjecture{}s. However, the model is still free to learn a different kind of bias which is an outcome of the fact that natural language explanations convey ideas through both content and form. If the form for explanations of different labels is discriminative, an unconstrained learning algorithm could learn to infer first the type of explanation and use it to infer the task. For example, given the input ($\mathrm{concat}_{ph}, t_x$), where $x \in $ \{entail, contradict, neutral\}, if a model could learn whether $t_x$ is an entailment explanation, it then only has to output whether $\mathrm{concat}_{ph}$ corresponds to an entailment relation. Essentially, high label accuracy can be achieved by inferring first what task to do using only the form of $t_x$.

\paragraph{The fix:} To prevent \system{} from exploiting the form of an \conjecture{} as described above, we create additional training examples, where we require \system{} to score valid instance-explanation pairs higher. In particular, we sample negative explanations for an instance, of the same form as the correct label. For example, an instance labeled as entailment would have an additional training signal: Score ($\mathrm{concat}_{ph}, t_\text{entail}$) higher than ($\mathrm{concat}_{ph}, t_\text{entail}'$) and ($\mathrm{concat}_{ph}, t_\text{entail}''$), where $t_\text{entail}'$ and $t_\text{entail}''$ are randomly sampled entailment form \conjecture{}s.

We note that the fix leaves room for other kinds of biases to be learnt. However, the key advantage with \system{} is that it is easy to design probes to test for such biases and subsequently fix them (see \refsec{sec:sensitivity_analysis}).

\subsection{Baselines}
\label{sec:baselines}
We now describe baselines which use the same underlying blocks as \system{}, for generating explanations and classification.

\paragraph{\system{}:post-hoc:} To understand the drop in performance which could be associated with constraining models as we have done, we train a model with full access to input examples (See \reffig{fig:f2_alternatives}A).
\begin{equation*}
	l_x = W_x F_\text{pre}(p, h)
\end{equation*}
where $x \in $ \{entail, contradict, neutral\}.

Further, we provide a strong baseline for post-hoc generators using this model, where using the model's predictions, we simply pick the corresponding label-specific generated \conjecture{}. 
\begin{equation*}
    t_g' =  G_\text{post}(l_x) = t_x
\end{equation*}
We note that the model's predictions have no sensitivity to the generated explanations in NILE: post-hoc.

\paragraph{ExplainThenPredictAttention (ETPA):} Following \cite{camburu2018snli}, (see \reffig{fig:f2_alternatives}B), we train a pipelined system, where we first learn to generate the gold explanation $t_g'$, followed by a classification of $t_g'$ to predict the label:
\begin{align*}
	t_g'   &= G_\text{pre}(\mathrm{concat}_{ecn}) \\
	l_x    &= W_x F_\text{post}(t_g')
\end{align*}
where $x \in $ \{entail, contradict, neutral\}.

\begin{table*}[thb]
\small
\centering
\begin{tabular}{|l|l|c|c|c|c|c|c|c|}
\hline
\multicolumn{2}{|c|}{\multirow{3}{*}{Model}} &
  \begin{tabular}[c]{@{}c@{}}MNLI\\ Dev\end{tabular} &
  \begin{tabular}[c]{@{}c@{}}MNLI\\ Dev-mm\end{tabular} &
  \multicolumn{5}{c|}{\begin{tabular}[c]{@{}c@{}}Explanation evaluation on \\ first 100 MNLI Dev Samples\end{tabular}} \\ \cline{3-9} 
\multicolumn{2}{|c|}{} &
  \multirow{2}{*}{\begin{tabular}[c]{@{}c@{}}Label\\ Accuracy\end{tabular}} &
  \multirow{2}{*}{\begin{tabular}[c]{@{}c@{}}Label\\ Accuracy\end{tabular}} &
  \multirow{2}{*}{\begin{tabular}[c]{@{}c@{}}A:\\ Correct\\ Labels \end{tabular}} &
  \multicolumn{2}{c|}{\begin{tabular}[c]{@{}c@{}}Averaged \\ over annotators\end{tabular}} &
  \multicolumn{2}{c|}{\begin{tabular}[c]{@{}c@{}}Annotators \\ in-agreement\end{tabular}} \\ \cline{6-9} 
\multicolumn{2}{|c|}{} &
   &
   &
   &
  \begin{tabular}[c]{@{}c@{}}B: \\ Correct\\ Expl.\end{tabular} &
  B/A &
  \begin{tabular}[c]{@{}c@{}}C:  \\ Correct\\ Expl.\end{tabular} &
  C/A \\ \hline \hline
ETPA \citep{camburu2018snli}                 & Reproduced  & 56.11 & 56.42 & 48 & 22.67 & 47.22 & 14 & 29.17 \\ \hline \hline
\multicolumn{2}{|l|}{NILE:post-hoc}    & \textbf{79.29} & \textbf{79.29} & 69 & 47.67 & 69.08 & 35 & 50.72 \\ \hline  \hline
\multirow{3}{*}{NILE-PH} & Independent & 54.95 & 55.35 & 46 & 34.33 & 74.64 & 28 & \textbf{60.87} \\ \cline{2-9} 
                         & Aggregate   & 56.45 & 56.66 & 49 & 34.67 & 70.75 & 26 & 53.06 \\ \cline{2-9} 
                         & Append      & 61.33 & 61.98 & 58 & 43.33 & \textbf{74.71} & 34 & 58.62 \\ \hline  \hline
\multirow{3}{*}{NILE-NS} & Independent & 74.84 & 75.20 & 68 & 49.67 & 73.04 & 37 & 54.41 \\ \cline{2-9} 
                         & Aggregate   & 75.73 & 76.22 & 69 & 49.33 & 71.50 & 37 & 53.62 \\ \cline{2-9} 
                         & Append      & 77.07 & 77.22 & \textbf{72} & \textbf{52.33} & 72.69 & \textbf{38} & 52.78 \\ \hline  \hline
\multirow{2}{*}{NILE}    & Independent & 72.91 & 73.04 & 64 & 45.67 & 71.35 & 33 & 51.56 \\ \cline{2-9} 
                         & Aggregate   & 72.94 & 73.01 & 63 & 45.67 & 72.49 & 34 & 53.97 \\ \hline
\end{tabular}
\caption{\label{tbl:label_accuracy_mnli} \textit{Testing the generalization capability of \system{} on the \textbf{out-of-domain} MNLI Dev sets.} Training and model selection is done on the SNLI dataset (\refsec{sec:overall_results}), and evaluation on the out-of-domain MNLI Dev (matched) and MNLI Dev-mm (mismatched) sets. Label accuracies are reported for both MNLI Dev (matched) and MNLI Dev-mm (mismatched) sets, while explanations are evaluated on the first 100 MNLI Dev set examples. We report the number of correct explanations, averaged across annotators (B) as well as when all annotators agree on correctness (C). All \system{} variants provide more correct explanations than the ETPA baseline (B, C). Further, the percentage of correct explanations in the subset of correct label predictions (B/A, C/A) is significantly better for all \system{} variants. The results demonstrate that \system{} provides a more generalizable framework for producing natural language explanations. Please see \refsec{sec:transfer_results} for details.}
\end{table*}

%% file: sections/experiments.tex
\section{Experiments}
\label{sec:expts}
In this section, we aim to answer the following questions:
\begin{description}[noitemsep,nolistsep]
\item[Q1] How does \system{} compare with the baselines and other existing approaches in terms of final task performance, and explanation accuracy, on in-domain evaluation sets (train and test on SNLI)? (\refsec{sec:overall_results})
\item[Q2] How well does \system{} transfer to out-of-domain examples (train on SNLI, and test on MNLI)? (\refsec{sec:transfer_results})
\item[Q3] How faithful are the model's predictions to the generated explanations? (\refsec{sec:sensitivity_analysis})

\end{description}
We provide training details in Appendix A, and examples of generated label-specific explanations in Appendix B.

\subsection{In-domain Results}
\label{sec:overall_results}
We report the label accuracies of the baselines and proposed architectures on the SNLI Dev and Test set in \reftbl{tbl:label_accuracy}. We also report explanation accuracies,  obtained through human evaluation of the generated explanations in the first 100 test examples. Binary scores on correctness were sought from five annotators (non-experts in NLP) on the generated explanations. For both label and explanation accuracies, we report using a model selected using the SNLI Dev set label accuracy across 5 runs with 5 different seeds of random initialization. Please see the Appendix for more details on the the 5 runs. First, through \system{}:post-hoc, we provide a strong baseline for obtaining high label and explanation accuracy. Our aim in this work is to learn explanations that serve as the reason for the model's predictions. Nevertheless, we are able to compete or outperform this baseline, in terms of explanation accuracy, while incurring a only a small drop in label accuracy. All variants of \system{}, including \nophsystem{} and \nonssystem{} (which is not trained using negative samples of explanations as described in \refsec{subsec:using_conjectures_and_examples}), produce more correct explanations than the ETPA baseline. \nophsystem{}:Append, \system{} and \nonssystem{} provide gains over label accuracies compared to the ETPA baseline. Additionally, \system{} and its variants provide natural ways to probe the sensitivity of the system's predictions to the explanations, as demonstrated in the subsequent sections. Finally, the explanations generated by all \system{} variants generalize significantly better on out-of-distribution examples when compared to the ETPA baseline (See \refsec{sec:transfer_results}).

\subsection{Transfer to Out-of-domain NLI}
\label{sec:transfer_results}
To test the generalization capability of \system{}, we do training and model selection on the SNLI dataset (\refsec{sec:overall_results}), and evaluate on the out-of-domain MNLI \citep{williams-etal-2018-broad} development sets. Transfer without fine-tuning to out-of-domain NLI has been a challenging task with transfer learning for generating explanations  in MNLI being particularly challenging \citep{camburu2018snli}. We report label accuracies on the Dev (matched) and Dev-mm (mismatched) sets, and explanation evaluation on the first 100 Dev samples in \reftbl{tbl:label_accuracy_mnli}. Explanation evaluation was done by three annotators (who also annotated the SNLI explanations).  While the label accuracies follow a similar pattern as the in-domain SNLI Test set, all variants of \system{} provide gains in the quality of generated explanations. All variants of \system{} produce more correct explanations (B, C) as well as a higher percentage of correct generated explanations among correct predictions (B/A, C/A). This demonstrates that \system{}, through intermediate label-specific natural language explanations, provides a more general way for building systems which can produce natural language explanations for their decisions.

\subsection{Evaluating Faithfulness using Sensitivity Analysis}
\label{sec:sensitivity_analysis}
\begin{table}[thb]
\small
\centering
\begin{tabular}{|l|l|c|c|c|}
\hline
\multicolumn{2}{|c|}{Model} & \begin{tabular}[c]{@{}c@{}}I+\\ Exp\end{tabular} & \begin{tabular}[c]{@{}c@{}}I\\ only\end{tabular} & \begin{tabular}[c]{@{}c@{}}Exp\\ only\end{tabular} \\ \hline
\multirow{3}{*}{\nonssystem{}} & Independent & 91.6 & 33.8 & 69.4 \\ \cline{2-5} 
 & Aggregate & 91.6 & 33.8 & 74.5 \\ \cline{2-5} 
 & Append & 91.7 & 91.2 & 72.9 \\ \hline
\multirow{2}{*}{\system{}} & Independent & 91.3 & 33.8 & 46.1 \\ \cline{2-5} 
 & Aggregate & 91.2 & 33.8 & 40.7 \\ \hline
\end{tabular}
\caption{\label{tbl:erasing} \textit{Estimating the sensitivity of the system's predictions to input explanations through erasure.} During testing, we erase either the instance or the explanations from the input to \nonssystem{} and \system{}. The results seem to indicate that \nonssystem{}'s predictions are more faithful, in the sense of having a higher sufficiency. However, as demonstrated subsequently, the sensitivity of \nonssystem{}'s prediction to the input explanations is not as desired. Please see \refsec{sec:sensitivity_analysis} for details.}
\end{table}

\begin{table}[thb]
\small
\centering
\begin{tabular}{|l|l|c|c|}
\hline
\multicolumn{2}{|c|}{Model} & Dev Set & \begin{tabular}[c]{@{}c@{}}Shuffled \\ Dev Set\end{tabular} \\ \hline
\multirow{3}{*}{\nonssystem{}} & Independent & 91.6 & 88.1 \\ \cline{2-4} 
 & Aggregate & 91.6 & 89.6 \\ \cline{2-4} 
 & Append & 91.7 & 88.5 \\ \hline
\multirow{2}{*}{\system{}} & Independent & 91.3 & 35.3 \\ \cline{2-4} 
 & Aggregate & 91.2 & 31.6 \\ \hline
\end{tabular}
\caption{\label{tbl:shuffle_test} \textit{Probing the sensitivity of the system's predictions by shuffling instance-explanation pairs.} Each instance is attached to a randomly selected explanation of the same form as the original pair. The results demonstrate a much weaker link between \nonssystem{}'s predictions and associated explanations. On the other hand, \system{} behaves more expectedly. Note that the baselines don't allow a similar mechanism to test their faithfulness, and such testability is a key advantage of \system{}. Please see \refsec{sec:sensitivity_analysis} for details.}
\end{table} 

\system{} and its variants allow a natural way to probe the sensitivity of their predictions to the generated explanations, which is by perturbing the explanations themselves. In this way, \system{} resembles explanation systems which provide input text fragments as reasons for their decisions. \citet{deyoung2019eraser} propose metrics to evaluate the faithfulness of such explanations. Following their work, we first attempt to measure the explanations generated by the methods proposed in this paper for comprehensiveness (what happens when we remove the explanation from the input) and sufficiency (what happens if we keep only the explanations). In \reftbl{tbl:erasing}, we show these measures for \system{} and \nonssystem{}. The results seem to indicate that explanations for both \system{} and \nonssystem{} are comprehensive, while having higher sufficiency in the case of \nonssystem{}. We first note that the comprehensiveness of these systems is ensured by design, and the input is indistinguishable without an explanation. Second, we argue that sufficiency may indicate correlations which don't necessarily exist in the system otherwise. We study the sensitivity of the explanations through a probe motivated by an understanding of the task and the training examples (see \refsec{subsec:using_conjectures_and_examples}). We perturb the instance-explanation inputs such that for each test instance, the explanation is replaced by a randomly selected explanation of the same label. The results (\reftbl{tbl:shuffle_test}) indicate that \nonssystem{} is more robust to random perturbations of input explanations, and presumably uses the form of the explanation to infer the task (see \refsec{subsec:using_conjectures_and_examples} for a discussion). It is true that \system{} behaves expectedly as we have specifically designed \system{} to prevent the associated bias, and that this could potentially lead the system to learn other such biases. However, a key advantage of the proposed architecture is the ability to identify and fix for such biases. We leave it as an interesting and challenging future work to find and fix more such biases.

%% file: sections/conclusion.tex
\section{Conclusion}
\label{sec:conclusion}

In this paper we propose \system{}, a system for Natural Language Inference (NLI) capable of generating labels along with natural language explanations for the predicted labels. Through extensive experiments, we demonstrate the effectiveness of this approach, in terms of both label and explanation accuracy. \system{} supports the hypothesis that accurate systems can produce testable natural language explanations of their decisions. In the paper, we also argue the importance of explicit evaluation of faithfulness of the generated explanations, i.e., how correlated are the explanations to the model's decision making. We evaluate faithfulness of \system{}'s explanations using sensitivity analysis. Finally, we demonstrate that task-specific probes are necessary to measure such sensitivity.

%% file: sections/appendix.tex
\section{Experimental Setup}
\label{sec:experimental_setup}

\begin{table}[hb]
\small
\centering
\begin{tabular}{|l|c|c|c|}
\hline
\multicolumn{2}{|c|}{\multirow{2}{*}{Model}} & \multicolumn{2}{c|}{\begin{tabular}[c]{@{}c@{}}SNLI Dev\\ Label Accuracy\end{tabular}} \\ \cline{3-4} 
\multicolumn{2}{|c|}{}                 & Mean  & Stddev \\ \hline
ETPA                     & Reproduced  & 86.96 & 0.02   \\ \hline
\multicolumn{2}{|l|}{NILE:post-hoc}    & 91.77 & 0.06   \\ \hline
\multirow{3}{*}{NILE-PH} & Independent & 84.53 & 0.18   \\ \cline{2-4} 
                         & Aggregate   & 85.47 & 0.26   \\ \cline{2-4} 
                         & Append      & 88.30 & 0.12   \\ \hline
\multirow{3}{*}{NILE-NS} & Independent & 90.17 & 2.76   \\ \cline{2-4} 
                         & Aggregate   & 91.44 & 0.06   \\ \cline{2-4} 
                         & Append      & 91.57 & 0.14   \\ \hline
\multirow{2}{*}{NILE}    & Independent & 91.09 & 0.19   \\ \cline{2-4} 
                         & Aggregate   & 90.94 & 0.22   \\ \hline
\end{tabular}
\caption{\label{tbl:snli_dev_mean_std} \textit{Mean and Standard Deviation for label accuracues on SNLI Dev set} are reported. \nonssystem{}:Independent system has a high standard deviation and relatively lower mean accuracy. This is due to a bad random initialization with seed 219. When seed 219 results are excluded, the mean and standard deviation are 91.41 and 0.20 respectively.}
\end{table}

For fine-tuning  gpt2-medium language models for explanation generation as well as roberta-base models, we leverage code and pre-trained models from the ``transformers" library available at {\UrlFont{https://github.com/huggingface}}. In each case we train on the train split for three epochs. Apart from batch size, sequence length and seed for random initialization, we keep the other hyperparameters fixed throughout the experiments. We don't do any fine-tuning on seeds of random initialization. For roberta-base models, we report results through model selection  on models trained using 5 seeds of random initialization - 42, 219, 291, 67 and 741. Model selection is done using label accuracies on SNLI Dev set. In \reftbl{tbl:snli_dev_mean_std}, we report the mean and standard deviation for the label accuracies across 5 runs.

We ran our experiments on GeForce GTX 1080 Ti GPUs. We adjust the batch size to be the largest multiple of 16 to fit on the GPU memory ($\sim$12GB). We now list all the hyper-parameters used. 

\paragraph{GPT2:} The hyper-parameters used for fine-tuning gpt2-medium include a maximum sequence length of 128, batch size of 2, learning rate of 5e-5, Adam epsilon of 1e-8, max gradient norm of 1.0 and a seed of 42. For generating text, we used greedy decoding.

\begin{table}[th]
\small
\centering
\begin{tabular}{|l|l|c|c|}
\hline
\multicolumn{2}{|c|}{Model} & \begin{tabular}[c]{@{}c@{}}Batch \\ size\end{tabular} & \begin{tabular}[c]{@{}c@{}}Max\\ seq \\ length\end{tabular} \\ \hline
ETPA                     & Reproduced  & 32 & 100   \\ \hline
\multicolumn{2}{|l|}{NILE:post-hoc}    & 32 & 100   \\ \hline
\multirow{3}{*}{\nophsystem{}} & Independent & 32 & 50 \\ \cline{2-4} 
 & Aggregate & 32 & 50 \\ \cline{2-4} 
 & Append & 32 & 100 \\ \hline
\multirow{3}{*}{\nonssystem{}} & Independent & 16 & 100 \\ \cline{2-4} 
 & Aggregate & 16 & 100 \\ \cline{2-4} 
 & Append & 16 & 200 \\ \hline
\multirow{2}{*}{\system{}} & Independent & 16 & 100 \\ \cline{2-4} 
 & Aggregate & 16 & 100 \\ \hline
\end{tabular}
\caption{\label{tbl:roberta_hyper_parameters} Hyper-parameters (batch size and maximum sequence length) used for fine-tuning roberta-base}
\end{table}

\paragraph{RoBERTa:} The fixed parameters for fine-tuning robera-base included a learning rate  of 2e-5, Adam epsilon of 1e-8 and max gradient norm 1.0. Other parameters are captured in \reftbl{tbl:roberta_hyper_parameters}.

\section{Generated Explanations}
\label{sec:generated_explanations}
In this section, we list the label-specific explanations generated for five Dev set examples, along with the premise, hypothesis and gold labels, for both SNLI (\refsec{subsec:snli_generated_explanations}) and MNLI (\refsec{subsec:mnli_generated_explanations}) datasets.

\subsection{SNLI}
\label{subsec:snli_generated_explanations}

\begin{itemize}[]
	\item \textbf{Example 1} \\
	\textbf{{PairID}}: 4705552913.jpg\#2r1n \\
	\textbf{{gold label}}: neutral \\
	\textbf{{Premise}}: Two women are embracing while holding to go packages. \\
	\textbf{{Hypothesis}}: The sisters are hugging goodbye while holding to go packages after just eating lunch. \\\\
	\textbf{{entailment}}: Sisters are women. \\
	\textbf{{contradiction}}: The women cannot be embracing if they are hugging goodbye. \\
	\textbf{{neutral}}: Two women embracing does not imply they are sisters hugging goodbye.
\end{itemize}
\begin{itemize}[]
	\item \textbf{Example 4} \\
	\textbf{{PairID}}: 2407214681.jpg\#0r1e \\
	\textbf{{gold label}}: entailment \\
	\textbf{{Premise}}: Two young children in blue jerseys, one with the number 9 and one with the number 2 are standing on wooden steps in a bathroom and washing their hands in a sink. \\
	\textbf{{Hypothesis}}: Two kids in numbered jerseys wash their hands. \\\\
	\textbf{{entailment}}: Two young children are two kids, and blue jerseys implies numbered jerseys. \\
	\textbf{{contradiction}}: The kids are either standing on wooden steps or wash their hands. \\
	\textbf{{neutral}}: The kids could be washing their hands in a sink for any reason, not just because they wash their hands.
\end{itemize}
\begin{itemize}[]
	\item \textbf{Example 18} \\
	\textbf{{PairID}}: 3667788497.jpg\#0r1c \\
	\textbf{{gold label}}: contradiction \\
	\textbf{{Premise}}: Under a blue sky with white clouds, a child reaches up to touch the propeller of a plane standing parked on a field of grass. \\
	\textbf{{Hypothesis}}: A child is playing with a ball. \\\\
	\textbf{{entailment}}: A propeller is a type of ball. \\
	\textbf{{contradiction}}: A child cannot touch the propeller of a plane while playing with a ball. \\
	\textbf{{neutral}}: Just because the child reaches up to touch the propeller of a plane does not mean the child is playing with a ball.
\end{itemize}
\begin{itemize}[]
	\item \textbf{Example 24} \\
	\textbf{{PairID}}: 6170429974.jpg\#3r1c \\
	\textbf{{gold label}}: contradiction \\
	\textbf{{Premise}}: Two men on bicycles competing in a race. \\
	\textbf{{Hypothesis}}: A few people are catching fish. \\\\
	\textbf{{entailment}}: Two men are a few people, and catching fish is a form of competition. \\
	\textbf{{contradiction}}: competing and catching are different actions \\
	\textbf{{neutral}}: Competing in a race does not imply catching fish.
\end{itemize}
\begin{itemize}[]
	\item \textbf{Example 30} \\
	\textbf{{PairID}}: 173499538.jpg\#2r1c \\
	\textbf{{gold label}}: contradiction \\
	\textbf{{Premise}}: Families waiting in line at an amusement park for their turn to ride. \\
	\textbf{{Hypothesis}}: People are waiting to see a movie. \\\\
	\textbf{{entailment}}: People are waiting to see a movie at an amusement park. \\
	\textbf{{contradiction}}: People cannot be waiting to ride an amusement park and a movie at the same time. \\
	\textbf{{neutral}}: Not all rides at amusement parks are a movie.
\end{itemize}

\subsection{MNLI}
\label{subsec:mnli_generated_explanations}

\begin{itemize}[]
	\item \textbf{Example 1} \\
	\textbf{{PairID}}: 63735n \\
	\textbf{{gold label}}: neutral \\
	\textbf{{Premise}}: The new rights are nice enough \\
	\textbf{{Hypothesis}}: Everyone really likes the newest benefits  \\\\
	\textbf{{entailment}}: Nice enough is another way of saying everyone really likes. \\
	\textbf{{contradiction}}: If everyone really likes theBenefits, then the rights wouldn't be nice enough. \\
	\textbf{{neutral}}: New benefits are not necessarily the newest benefits.
\end{itemize}
\begin{itemize}[]
	\item \textbf{Example 14} \\
	\textbf{{PairID}}: 73734e \\
	\textbf{{gold label}}: entailment \\
	\textbf{{Premise}}: You and your friends are not welcome here, said Severn. \\
	\textbf{{Hypothesis}}: Severn said the people were not welcome there. \\\\
	\textbf{{entailment}}: Friends are people. \\
	\textbf{{contradiction}}: People who are not welcome here are not welcome here. \\
	\textbf{{neutral}}: You can not infer they were not welcome.
\end{itemize}
\begin{itemize}[]
	\item \textbf{Example 17} \\
	\textbf{{PairID}}: 70726e \\
	\textbf{{gold label}}: entailment \\
	\textbf{{Premise}}: so i have to find a way to supplement that \\
	\textbf{{Hypothesis}}: I need a way to add something extra. \\\\
	\textbf{{entailment}}: Add something extra is a rephrasing of Supplement that. \\
	\textbf{{contradiction}}: A way to supplement something extra is not the same as a way to add something extra. \\
	\textbf{{neutral}}: Just because someone is adding something doesn't mean they are doing it extra.
\end{itemize}
\begin{itemize}[]
	\item \textbf{Example 26} \\
	\textbf{{PairID}}: 67610c \\
	\textbf{{gold label}}: contradiction \\
	\textbf{{Premise}}: Sorry but that's how it is. \\
	\textbf{{Hypothesis}}: This is how things are and there are no apologies about it. \\\\
	\textbf{{entailment}}: oops that's how it is is is same as there are no apologies about it \\
	\textbf{{contradiction}}: A person can't be sorry and have no apologies. \\
	\textbf{{neutral}}: Just because someone is sorry does not mean they are saying no apologies.
\end{itemize}
\begin{itemize}[]
	\item \textbf{Example 45} \\
	\textbf{{PairID}}: 98811c \\
	\textbf{{gold label}}: contradiction \\
	\textbf{{Premise}}: yeah i mean just when uh the they military paid for her education \\
	\textbf{{Hypothesis}}: The military didn't pay for her education. \\\\
	\textbf{{entailment}}: The military paid for her education, doesn't matter if it was for college or not. \\
	\textbf{{contradiction}}: The military either paid for her education or they didn't. \\
	\textbf{{neutral}}: Just because the military paid for her education doesn't mean she didn't get paid for it.
\end{itemize}